\documentclass[sigconf,screen]{acmart}
\AtBeginDocument{%
  \providecommand\BibTeX{{%
    \normalfont B\kern-0.5em{\scshape i\kern-0.25em b}\kern-0.8em\TeX}}}

\setcopyright{acmlicensed}
\copyrightyear{2024}
\acmYear{2024}
\acmDOI{XXXXXXX.XXXXXXX}

\acmConference[HRI '24 Workshop]{Workshop of the 2024 ACM/IEEE International Conference on Human-Robot Interaction}{March 11--14, 2024}{Boulder, CO, USA}
\acmBooktitle{Workshop of the 2024 ACM/IEEE International Conference on Human-Robot Interaction (HRI '24 Workshop), March 11--14, 2024, Boulder, CO, USA}
\acmISBN{978-1-4503-XXXX-X/24/06}

\begin{document}

\title{Multimodal Human-Autonomous Agents Interaction Using Pre-Trained Language and Visual Foundation Models}

\author{Linus Nwankwo}
\email{linus.nwankwo@unileoben.ac.at}
\affiliation{%
  \institution{Chair of Cyber-Physical Systems, Montanuniversität}
  \streetaddress{Franz Josef-Straße 18, 8700 Leoben}
  \city{Leoben}
  \country{Austria}
  \postcode{8700}
}

\author{Elmar Rueckert}
\affiliation{%
  \institution{Chair of Cyber-Physical Systems, Montanuniversität}
  \streetaddress{Franz Josef-Straße 18, 8700 Leoben}
  \city{Leoben}
  \country{Austria}}


\renewcommand{\shortauthors}{Linus Nwankwo and Elmar Rueckert}
\begin{abstract}
  In this paper, we extended the method proposed in \cite{linushri2024} to enable humans to interact naturally with autonomous agents through vocal and textual conversations.  Our extended method exploits the inherent capabilities of pre-trained large language models (LLMs), multimodal visual language models (VLMs), and speech recognition (SR) models to decode the high-level natural language conversations and semantic understanding of the robot's task environment, and abstract them to the robot's actionable commands or queries. We performed a quantitative evaluation of our framework's natural vocal conversation understanding with participants from different racial backgrounds and English language accents. The participants interacted with the robot using both spoken and textual instructional commands. Based on the logged interaction data, our framework achieved \(87.55\%\) vocal commands decoding accuracy, \(86.27\%\) commands execution success, and an average latency of \(0.89\) seconds from receiving the participants' vocal chat commands to initiating the robot’s actual physical action. The video demonstrations of this paper can be found at \url{https://linusnep.github.io/MTCC-IRoNL/}.
\end{abstract}





\maketitle

\section{Introduction}\label{sec:intro}
Existing approaches for interacting with autonomous robots in the real world have been dominated by complex teleoperation controllers \cite{teleop} and rigid command protocols \cite{protocol}, where the robots execute predefined tasks based on specialized programming languages. As the challenges we present to these robots become more intricate and the environments they operate in grow more unpredictable \cite{linus}, there arises an unmistakable need for more natural and intuitive interaction mechanisms. 

The last few years have witnessed tremendous advancement in generative AI and natural language processing (NLP) \cite{bengesi2023advancements}, \cite{genAI}. These advancements driven primarily by foundation models, specifically transformer-based large language models (LLMs) like OpenAI GPT-3 \cite{gpt3}, Google BERT \cite{bert}, Meta AI LLaMA \cite{llama}, and multimodal visual language models (VLMs) e.g., CLIP \cite{clip}, DALL-E \cite{dalle}, and their successors, has opened new avenues for human-robot interaction (HRI) \cite{hri}. The inherent abilities of these models to understand language patterns, and structure and generate human-like responses as well as visual observations have led to several interesting robotic applications, such as \cite{brohan2023rt2}, \cite{brohan2023can} and \cite{lang1}.

In this paper, we exploit the inherent natural language capabilities of the pre-trained foundation models, as well as a speech recognition (SR) model to enable humans to interact naturally with autonomous agents through both spoken and textual dialogues.
As demonstrated in the video at the project website\footnote{\url{https://linusnep.github.io/MTCC-IRoNL/}}, our framework aims to realize a new approach to human-robot interactions—one where the vocal or textual conversation is the command (refer to Section \ref{sec:method} for more details).
Therefore, our contributions are twofold:
\begin{itemize}
\item we introduced a dual-modality framework that can leverage independent pre-trained LLMs, VLMs, and SR models to enable humans to interact with real-world autonomous robots or other entities through spoken or textual conversations. 
\item we performed real-world experiments with our developed framework to ensure that the robot's actions are always aligned with the user's spoken or textual instructions. 
\end{itemize}

\section{Related Work}
Prior works such as \cite{ferrari2023facilitating}, \cite{MARGE2022101255}, and \cite{10.1145/3568294.3580053} have explored the incorporation of vocal instructions into robotic systems. However, while these works are exceptional, they have relied primarily on a direct speech-to-action (STA) approach, where the robot's actions are dependent upon the accurate transcription of the vocal commands by the SR model employed in the respective works. In most noisy real-world scenarios, their approach may introduce stochastic behaviour in the robot's actions due to vulnerability to acoustic distortions present in real-world environments. 

Further, L. Nwankwo et al. \cite{linushri2024} incorporated text-based interaction techniques into autonomous robots leveraging LLMs and VLMs. Nonetheless, the framework lacks complete naturalness due to the absence of a mechanism to understand vocal instructions. In this work, we build upon the foundation provided in \cite{linushri2024}. Instead of relying on the accuracy of the SR model to plan the robot’s actions or depend on the text-based approach as a standalone, we propose a dual-modality approach that synergizes both the textual and vocal modalities. We leveraged the LLMs and the SR models' abilities to maintain robustness in diverse environments. In environments where ambient noise levels may compromise the accuracy of the SR model's vocal instructions decoding, our framework provides the flexibility to revert to the text-based interaction method. Conversely, in a quieter environment, the user can leverage the vocal modality pipeline for more natural and seamless interaction.

With our proposed dual-modality approach, we aim to provide the user with the autonomy to select the mode of interaction most suited to the prevailing conditions. Specifically, our framework mitigates the risk of misinterpretations and erroneous robotic actions that may have arisen due to sole dependence on the STA method, thereby ensuring consistent and reliable HRI \cite{hri} in the real world.

\section{Method}\label{sec:method}
\begin{figure}[h]
  \centering
  \includegraphics[width=\linewidth]{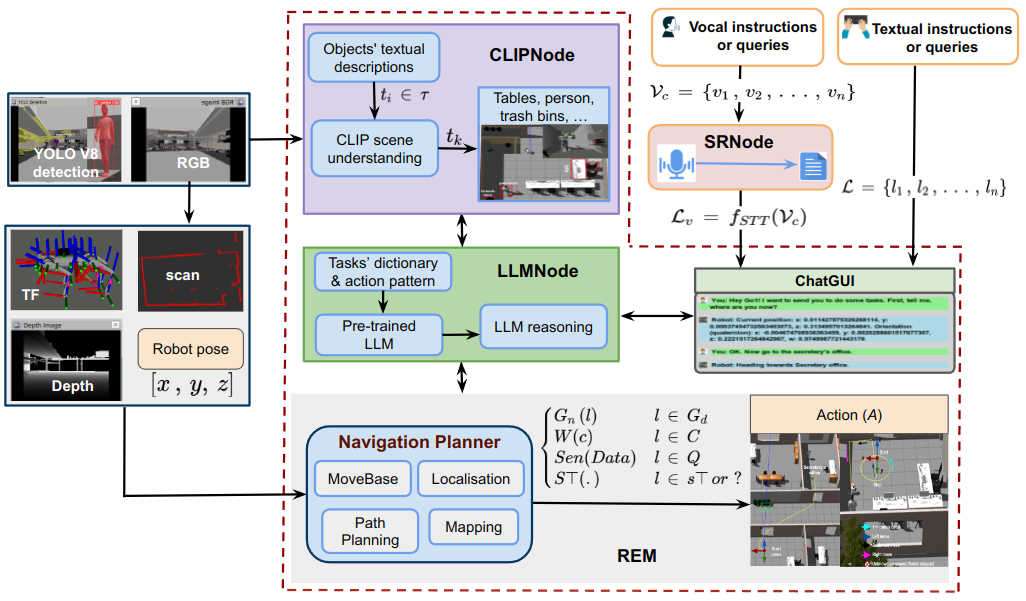}
  \caption{Overview of our framework's architecture. The area enclosed with the red dotted line decodes the textual-based natural language conversations and visual understanding. In the SRNode, we employed Google's SR model \cite{stt} to decode the vocal conversation from humans and abstract them to the textual representations required by the ChatGUI to interact with the LLMNode.}
  \label{fig:method}
\end{figure}

Figure \ref{fig:method} shows the architectural overview of our proposed framework.
The proposed model contains five main components: the LLMNode to decode the high-level textual-based conversations from humans, the CLIPNode to provide a visual and semantic understanding of the robot's task environment, the REM node to abstract the high-level understanding from the LLMNode to actual robot actions, the ChatGUI to serve as the user’s primary textual-based interaction point, and the SRNode to provide vocal or auditory commands understanding.

In this section, we provide details on the incorporation of the vocal conversation understanding pipeline. For details about the implementation of the first four sections (the area enclosed in red dotted line in Figure \ref{fig:method}) and how the pre-trained LLMs and VLMs are prompted to generate the actions used by the REM node, we refer the reader to \cite{linushri2024}.

\subsection{Vocal Conversation Decoding}\label{subsec:stt}
In order to decode the vocal natural language conversation and abstract them to the robot’s actions, we developed the SRNode. The SRNode employs Google's SR model \cite{stt} to capture the high-level auditory input from a microphone device, transcribing the auditory inputs to textual representations. The textual representation is subsequently used by the ChatGUI to establish communication between the LLMNode and the rest of the interfaces of Figure \ref{fig:method} within ROS ecosystem \cite{ros}.
Formally, given a vocal command \(\mathcal{V}_c\) e.g., task descriptions, queries captured by the microphone device, we developed a function \(f_{STT}\) employing the Google's SR model \cite{stt} such that \(\mathcal{V}_{c} = \{v_{1}, v_{2}, . . ., v_{n}\}\) is transcribed to a textual natural language representation \(\mathcal{L}\) as depicted in Eq. \ref{eqn:31}. The elements \(v_{i}\), \(i \in n\) represent distinct vocal commands or instructions given to the SR model through speech, e.g., ``Hello robot, can you move forward?'', ``What is your current location?'', ``Navigate to the kitchen area'', etc.
\begin{equation}\label{eqn:31}
    \mathcal{L}_{v} = f_{STT} (\mathcal{V}_{c}) = \{l_{1}, l_{2}, . . ., l_{n}\}, \quad l_{i} \in \mathcal{L}
\end{equation}
where \(l_{i}\) denotes the transcribed natural language command from the set \(\mathcal{L}_{v}\). We sent the resulting output from Eq.\ref{eqn:31} as input to the ChatGUI. We then used the LLMNode to handle the incoming natural language inputs from the ChatGUI by first passing them through the pre-trained LLM \cite{gpt2}. The resultant output from the LLMNode is then mapped to the robot's actionable commands or information request by the robot's execution mechanism (REM) node, consisting of the ROS\cite{ros} navigation planner packages 
shown at the lower bottom-left of Figure \ref{fig:method}.

\section{Experiments}
We conducted both real-world and simulated experiments to validate the performance of our framework. In simulation, we utilised the Unitree Go1 ROS \& Gazebo packages\footnote{\url{https://github.com/unitreerobotics/unitree_guide}} 
and a ROS-based open-source mobile robot adapted from \cite{NWANKWO2023e00426}. We ran all the simulations with a ground station PC with Nvidia Geforce RTX 3060 Ti GPU, 8GB memory running Ubuntu 20.04, ROS Noetic distribution.

In the real-world experiments, we used a Lenovo ThinkBook Intel Core i7 with Intel iRIS Graphics running  Ubuntu 20.04, ROS Noetic distribution. Segway RMP Lite 220 mobile robot was used. The robot is equipped with an RGB-D camera and a \(2D ~\text{RP}\)LiDAR for both visual and spatial observations of the task environment. We used our PC's inbuilt microphone and a plug-and-play AmazonBasics Pro Gaming Headset with a microphone function in all the experiments. We performed all the real-world experiments in our laboratory office (11 rooms) and outside corridor environment, measuring approximately \(18 \times 20\;m\) and \(6 \times 120\;m\) respectively.

We experimented with OpenAI GPT-2 \cite{gpt2}, Google BERT \cite{bert}, and Meta AI LLaMA \cite{llama}. OpenAI GPT-3 \cite{gpt3} and GPT-4 \cite{gpt4} are also adaptable to our framework. However, due to their API access limitations, we mostly utilised the open-access and free versions of the LLMs (GPT-2 \cite{gpt2} specifically) in our experiments. 

\subsection{Preliminary Results}
We invited \(5\) participants (average age of \(27 \ (\pm 3)\) and gender distribution, \(80\%\) male and \(20\%\) female) with different English language accents to interact with the robot via natural vocal conversation. We logged the interaction data i.e., the SR models' transcription of the participants' spoken words, the LLMNode predicted labels, the true action labels, etc. 
We used the logged interaction data to quantitatively evaluate the performance of our framework. 
We defined vocal commands understanding accuracy (VCUA) metric in addition to the navigation success rate (NSR), object identification accuracy (OIA), and average response time (ART) metrics utilised in \cite{linushri2024} to assess our framework's performance. With the VCUA, we assess how accurately the LLMNode predicts the commands based on the transcribed vocal instructions from the SRNode. We computed the accuracy as the percentage proportion of the correctly transcribed instruction to the generated instructions (from LLMNode) fed to the REM node for the actual robot's execution.

\begin{figure}[h]
    \centering
    \includegraphics[width=\linewidth]{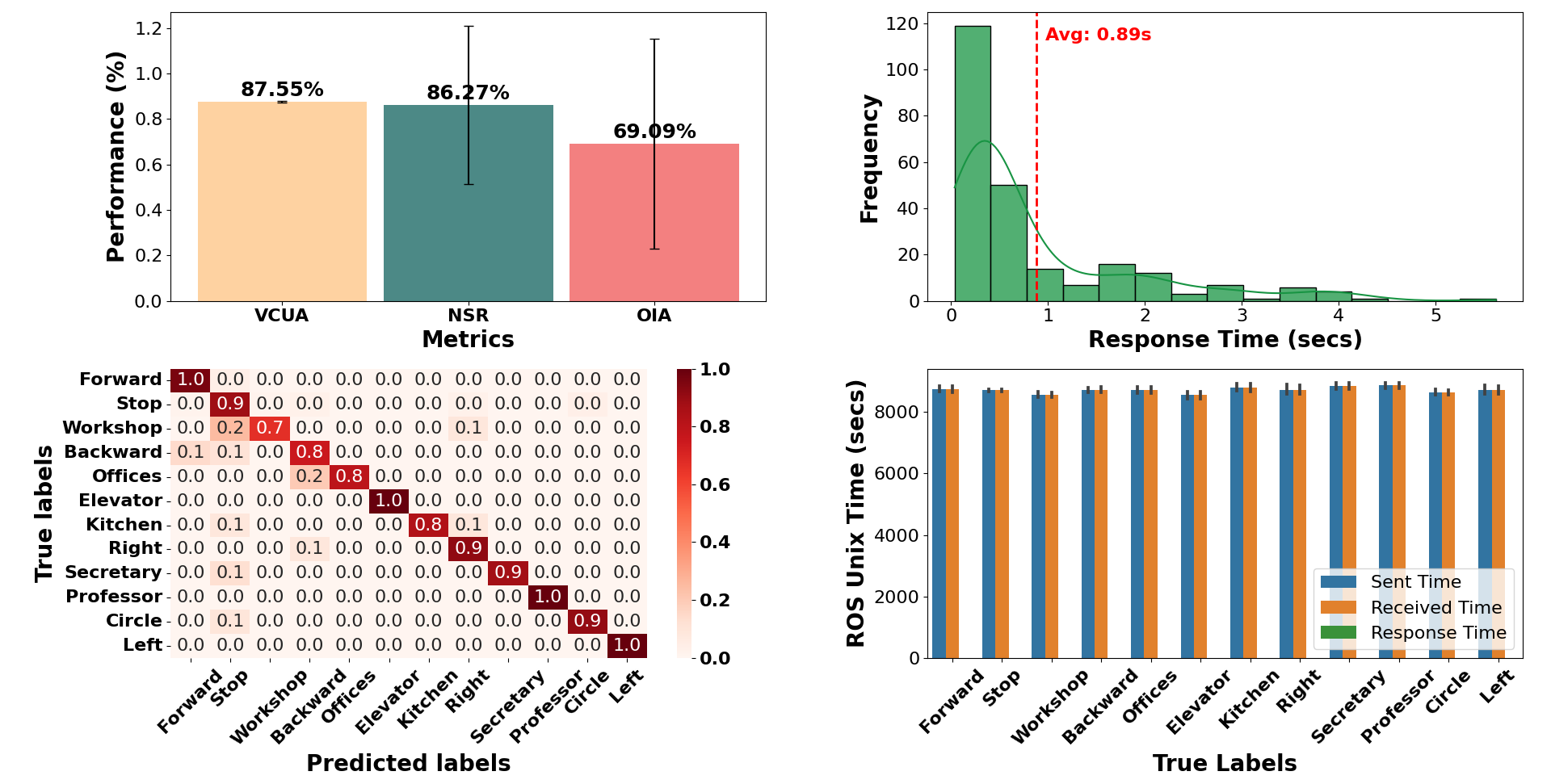}
    \caption{Quantitative evaluation results illustrating VCUA, NSR, OIA, and ART based on the logged interaction data.}
    \label{fig:metrics}
\end{figure}

Figure \ref{fig:metrics} presents the statistical results that we obtained from the interaction data analysis.
\vspace*{-0.90\baselineskip}
The top-left figure shows the VCUA, NSR, and OIA metrics for selected labels. We achieved \(87.55\%\) VCUA and \(86.27\%\) NSR, which indicates a good level of accuracy in the vocal commands decoding. In comparison to the results obtained in the textual-based method \cite{linushri2024}, we observed a slight difference in the command recognition accuracy, with VCUA achieving about \(12\%\) less than the textual-based CRA (\(99.13\%\)) as well as a \(11.69\%\) reduction in the NSR. This is expected because of the ambient environmental noise and variation in the participant's accents, which affect the vocal transcription from the SRNode, as could be seen in the confusion matrix (bottom-left of Figure \ref{fig:metrics}).

Further, the ART (right column of Figure \ref{fig:metrics}) across all the selected commands is approximately \(0.89\) seconds. This indicates that, on average, the robot takes less than a second from receiving a vocal chat command to initiating the robot's actual physical action, which suggests a relatively quick response time for our framework.

\section{Conclusion and Future Work}
We introduced a framework that leverages the inherent capabilities of LLMs, VLMs, and SR models to enhance human-robot interaction through natural vocal and textual conversations. Our evaluation from logged human interaction data achieved high vocal command understanding accuracy and effective task execution. This shows that our framework can enhance the intuitiveness and naturalness of human-robot interaction in the real world.
In our future work, we aim to refine our framework to resist the impact of environmental noise. We intend to incorporate adaptive noise-cancellation algorithms and context-aware speech recognition techniques to mitigate the impacts of random noise.
\vspace*{-0.50\baselineskip}

\bibliographystyle{ACM-Reference-Format}
\bibliography{sample-base}

\appendix



\end{document}